\documentclass[letterpaper]{article} % DO NOT CHANGE THIS
\usepackage[preprint]{aaai2027}
\usepackage[hyphens]{url}  % DO NOT CHANGE THIS
\usepackage{graphicx} % DO NOT CHANGE THIS
\usepackage{natbib}  % DO NOT CHANGE THIS AND DO NOT ADD ANY OPTIONS TO IT
\usepackage{caption} % DO NOT CHANGE THIS AND DO NOT ADD ANY OPTIONS TO IT
\usepackage{algorithm}
\usepackage{algorithmic}
\usepackage{amsmath}
\usepackage{amssymb}
\usepackage{newfloat}
\usepackage{listings}
\DeclareCaptionStyle{ruled}{labelfont=normalfont,labelsep=colon,strut=off} % DO NOT CHANGE THIS
\floatstyle{ruled}
\newfloat{listing}{tb}{lst}{}
\floatname{listing}{Listing}

\usepackage{booktabs}

\title{PIF-3D: Segmentation-Free Point-Image Fusion for Fine-Grained Text-to-3D Retrieval}

\author{
    Sankalp Sinha\textsuperscript{\rm 1 \rm 2}
    Didier Stricker\textsuperscript{ \rm 1 \rm 2}
    Muhammad Zeshan Afzal \textsuperscript{\rm 1 \rm 2}
}
\affiliations{
    \textsuperscript{\rm 1}German Research Center for Artificial Intelligence (DFKI), Germany\\
    \textsuperscript{\rm 2}RPTU Kaiserslautern-Landau, Germany\\
    \{sankalp.sinha, didier.stricker, muhammad\_zeshan.afzal\}@dfki.de
}

\begin{document}
\maketitle

%%%%%%%%%%%%%%%%%%%%%%%%%%%%%%%%%%%%%%%%%%%%%%%%%%%% Abstract %%%%%%%%%%%%%%%%%%%%%%%%%%%%%%%%%%%%%%%%%%%%%%%%%%%%
\begin{abstract}
Fine-grained text-to-3D retrieval depends on cues distributed across geometry and appearance. Point clouds preserve three-dimensional structure, while rendered views expose complementary visual evidence and strong vision-language priors. Yet independently pooling these modalities prevents them from interacting before retrieval. We introduce \textbf{PIF-3D}, a segmentation-free framework that learns a joint point-image representation using only shape-level correspondences. Its PIF-Joint Transformer (PIF-JT) contextualizes local PointBERT tokens together with pooled multi-view tokens from a locked dino.txt image encoder, then attention-pools them into a unified shape embedding. Low-rank adaptation specializes the dino.txt text branch, and symmetric contrastive learning aligns shapes with captions. PIF-3D requires no part labels, generated captions, or auxiliary 3D--text pretraining. On Text2Shape, it establishes a new state of the art across all six metrics, achieving $32.16\pm1.09$ shape-to-text RR@1 and $18.86\pm0.19$ text-to-shape RR@1. We further use 3DCoMPaT200 to test whether this representation can satisfy multiple localized part-level constraints simultaneously. Under a controlled public-release adaptation, PIF-3D achieves $26.50\pm0.28$ R@1 and $79.75\pm0.52$ R@5 on six-part queries, exceeding the strongest non-PIF baseline by $5.05$ and $7.49$ points. These gains, together with our ablations, show that joint point-image modeling is particularly valuable as queries become more compositional. 
% Our code and experimental artifacts will be made publicly available.
\end{abstract}

%%%%%%%%%%%%%%%%%%%%%%%%%%%%%%%%%%%%%%%%%%%%%%%%%%%% Introduction %%%%%%%%%%%%%%%%%%%%%%%%%%%%%%%%%%%%%%%%%%%%%%%%%%%%
\section{Introduction}
\label{sec:introduction}
Language provides a natural interface for searching large 3D asset collections, a longstanding problem in 3D shape retrieval~\cite{tangelder2008survey,li2023crossmodalreview}. Unlike category labels, it can express fine-grained requirements such as curved arms, a circular base, or a red seat on a dark wooden chair. Retrieving the correct asset therefore requires matching language to local geometry, appearance, and part-level attributes. Text2Shape evaluates such detailed shape descriptions~\cite{chen2018text2shape}, while 3DCoMPaT200 tests compositional queries containing one, three, or six part--material--color constraints~\cite{ahmed20253dcompat200}.

The evidence needed for this task is distributed across complementary 3D representations~\cite{xiao2020deepgeometry}. Colored point clouds preserve object-centered geometry and surface samples~\cite{guo2021pointcloudsurvey}, while rendered views expose silhouettes, materials, and appearance cues through mature 2D visual encoders~\cite{alzahrani2024multiviewsurvey}. They are also directly compatible with pretrained vision-language models~\cite{gan2022vlpsurvey}. Multimodal-learning surveys identify cross-modal interactions as central to representation learning~\cite{baltrusaitis2019multimodal,liang2024multimodalsurvey}. Compressing point clouds and rendered views independently into global descriptors can therefore discard useful interactions between geometric regions and visual evidence before retrieval.

Previous methods have improved text-to-shape retrieval through stronger representations and additional supervision. Early approaches use colored voxels or sequences of rendered views~\cite{chen2018text2shape,han2019y2seq2seq}, while TriCoLo aligns text, images, and voxels through trimodal contrastive learning~\cite{ruan2024tricolo}. Parts2Words introduces part-word alignment using semantic part supervision~\cite{tang2023parts2words}, COM3D models cross-view correspondence~\cite{wu2024com3d}, and SCA3D generates additional training pairs from segmented components~\cite{ren2025sca3d}. Uni3D and 3DAlign-DAER instead benefit from large-scale multimodal 3D pretraining~\cite{zhou2024uni3d,fan20263dalign}. These advances demonstrate the value of local structure, rendered appearance, and additional data, but raise a complementary question: can fine-grained retrieval be learned from ordinary shape-level correspondences, without part supervision, generated captions, or external 3D--text data?

We answer this question with \textbf{PIF-3D}, a segmentation-free Point-Image Fusion framework. Each asset is represented by a colored point cloud and multiple rendered views. A trainable PointBERT encoder extracts local point-group tokens~\cite{yu2022pointbert}, while a locked dino.txt image encoder extracts language-aligned patch features from each rendering~\cite{jose2025dinov2}. The projected patches are mean-pooled within each view, producing one token per rendering. The PIF-Joint Transformer (PIF-JT) concatenates these view tokens with the point tokens and contextualizes them using an $L$-layer Transformer before learned attention pooling. Geometry and multi-view appearance can therefore interact before being compressed into the final shape embedding.

We align the shape embedding with captions using a symmetric contrastive objective. Following LiT, which demonstrates the effectiveness of adapting language to a locked pretrained image encoder~\cite{zhai2022lit}, we freeze the dino.txt image encoder and specialize the final text Transformer blocks using LoRA~\cite{hu2022lora}. PIF-3D thus exploits a large-scale 2D vision-language prior without external 3D--text pretraining, part annotations, point cloud segmentation labels, or generated captions.

On Text2Shape, PIF-3D establishes a new state of the art across all six reported metrics. Over three seeds, it achieves $32.16\pm1.09$ S2T RR@1 and $18.86\pm0.19$ T2S RR@1, exceeding the strongest prior results by $4.05$ and $1.33$ points. Ablations show that joint fusion outperforms the point-only and image-only variants as well as matched-input late fusion. We also conduct a controlled public-release evaluation on 3DCoMPaT200, retraining all compared systems with the same training data, global batch size, and number of optimizer updates. PIF-3D provides its largest gains on the most compositional six-part queries, reaching $26.50\pm0.28$ R@1 and $79.75\pm0.52$ R@5.

Our contributions are threefold:
\begin{itemize}
    \item We introduce a segmentation-free framework for fine-grained text-to-3D retrieval that requires neither part-level supervision nor additional 3D--text data.
    \item We propose the PIF-Joint Transformer (PIF-JT), which jointly contextualizes local point-group tokens and pooled multi-view tokens before global compression.
    \item We establish state-of-the-art results on Text2Shape and strong compositional retrieval performance under a controlled public-release adaptation of 3DCoMPaT200, supported by modality, fusion, and language-adaptation ablations.
\end{itemize}

%%%%%%%%%%%%%%%%%%%%%%%%%%%%%%%%%%%%%%%%%%%%%%%%%%%% Related Work %%%%%%%%%%%%%%%%%%%%%%%%%%%%%%%%%%%%%%%%%%%%%%%%%%%%
\section{Related Work}
\label{sec:related_work}

\subsection{Text-to-Shape Retrieval}
Text2Shape established a benchmark for fine-grained text-to-shape retrieval by collecting natural-language descriptions of colored ShapeNet chairs and tables and learning a joint embedding of text and voxelized shapes~\cite{chen2018text2shape}. Y2Seq2Seq replaced dense voxels with sequences of rendered views and learned cross-modal representations through coupled reconstruction and prediction of view and word sequences~\cite{han2019y2seq2seq}. More recently, TriCoLo jointly aligned text, multi-view images, and colored voxels using a trimodal contrastive objective~\cite{ruan2024tricolo}. These methods demonstrate the value of rendered appearance and contrastive learning, but encode each modality into an independent global representation.

PIF-3D instead jointly contextualizes local point-group tokens and pooled view tokens before constructing the shape embedding. Beyond whole-shape descriptions, 3DCoMPaT200 introduced Compositional Part Shape Retrieval, where queries specify one, three, or six part--material attributes~\cite{ahmed20253dcompat200}. We use this benchmark to evaluate whether PIF-3D transfers from fine-grained shape descriptions to controlled compositional queries.

\subsection{Fine-Grained Matching with Parts and Views}
Parts2Words directly represents shapes as point clouds and uses optimal transport to align semantic part embeddings with contextualized word embeddings~\cite{tang2023parts2words}. Although effective for local correspondence, it requires semantic part labels to train its segmentation pathway. COM3D instead enriches PointNet++~\cite{qi2017pointnetplusplus} features with multi-view representations extracted by a Scene Representation Transformer using rendered images, camera poses, and rays~\cite{wu2024com3d}. Its fused descriptors are matched to text using Earth Mover's Distance, cosine similarity, and semi-hard negative mining.

SCA3D addresses limited 3D--text data through online augmentation~\cite{ren2025sca3d}. It uses PartNet segmentations and an MLLM to construct a library of captioned components, then assembles new shapes and their corresponding captions during training. These approaches improve fine-grained retrieval through either part supervision or generated training pairs. In contrast, PIF-3D uses only the benchmark-provided shape-level correspondences, without semantic part labels, segmentation masks, generated captions, or augmented 3D--text pairs. Its PIF-JT learns interactions between local point-group tokens and pooled multi-view tokens without explicit part decomposition.

\subsection{Multimodal Pretraining and Visual Priors}
CLIP demonstrated that large-scale contrastive image--text pretraining produces transferable cross-modal representations~\cite{radford2021learning}. ULIP transfers such priors to 3D by aligning point clouds with the image and text representations of a pretrained vision-language model~\cite{xue2023ulip}. OpenShape scales this paradigm to heterogeneous 3D collections while filtering and enriching noisy textual descriptions~\cite{liu2023openshape}. Uni3D further scales 3D representation learning through a 2D-initialized ViT point encoder trained to match image--text-aligned teacher features~\cite{zhou2024uni3d}. These methods produce broadly transferable 3D representations but rely on auxiliary multimodal 3D pretraining.

For fine-grained alignment at scale, 3DAlign-DAER constructs Align3D-2M and learns hierarchical token-to-point attention refined by a dynamic attention policy, together with a hierarchical retrieval strategy~\cite{fan20263dalign}. Its strong alignment is supported by approximately two million curated 3D--text pairs. PIF-3D instead targets benchmark-supervised retrieval: it inherits a large-scale \emph{2D} vision-language prior but introduces no auxiliary 3D--text pretraining data.

Specifically, PIF-3D builds on dino.txt, which aligns a frozen DINOv2 visual backbone with language while preserving features suitable for both global and dense prediction~\cite{jose2025dinov2}. PIF-3D combines these language-aligned visual features with a trainable PointBERT encoder~\cite{yu2022pointbert}; projected patch features are pooled within each view and then jointly contextualized with point-group tokens. The pretrained image tower remains locked, while low-rank adaptation specializes the text branch for fine-grained 3D descriptions~\cite{hu2022lora}.

%%%%%%%%%%%%%%%%%%%%%%%%%%%%%%%%%%%%%%%%%%%%%%%%%%%% Method %%%%%%%%%%%%%%%%%%%%%%%%%%%%%%%%%%%%%%%%%%%%%%%%%%%%
\section{Method}
\label{sec:methodology}

\begin{figure*}[t]
    \centering
    \includegraphics[
        width=\linewidth,
        keepaspectratio
    ]{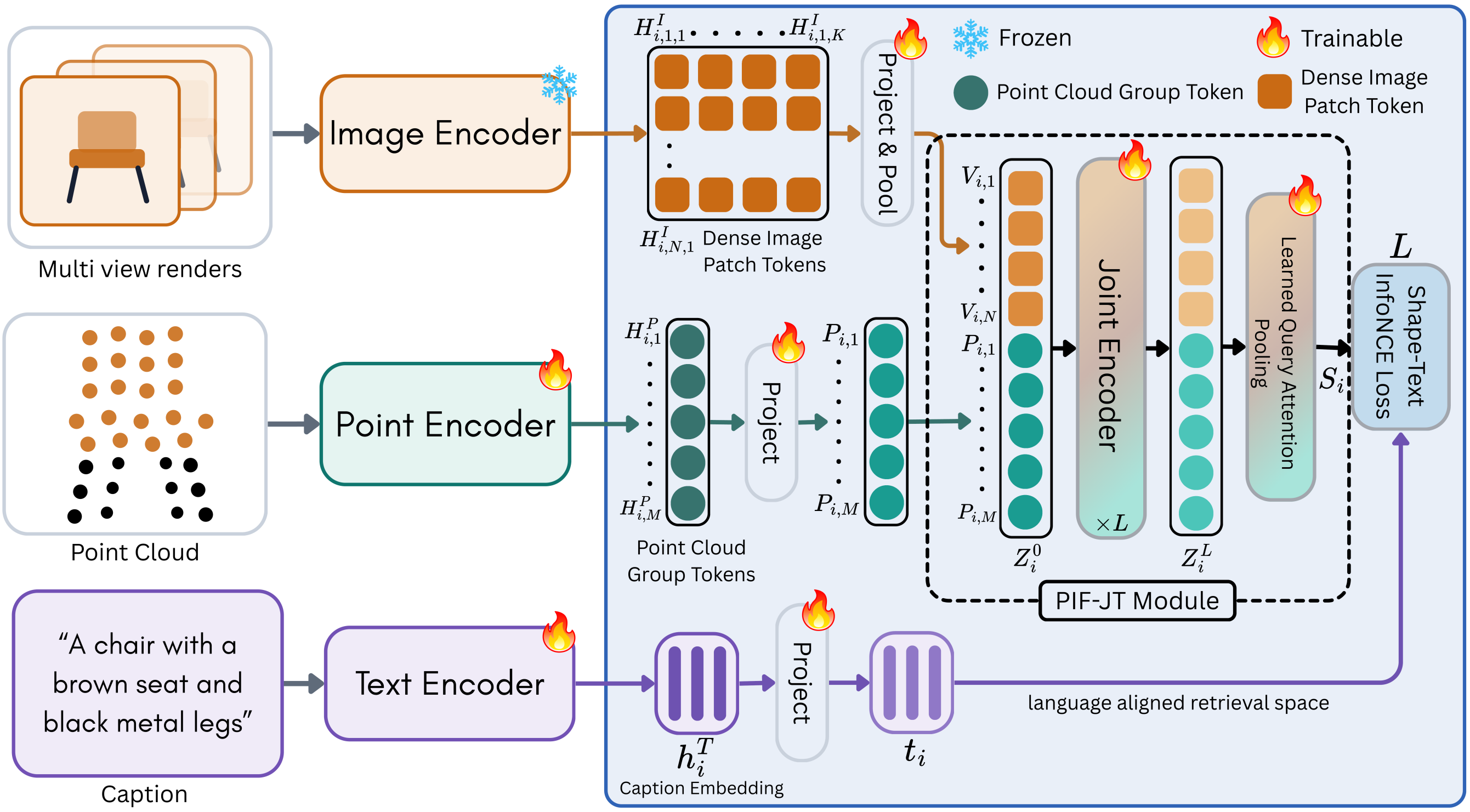}
    \caption{Overview of PIF-3D. PIF-JT jointly contextualizes projected PointBERT group tokens with language-aligned view tokens from the locked dino.txt image encoder, then attention-pools them into a shape embedding. Contrastive alignment with a LoRA-adapted dino.txt text embedding preserves the pretrained visual space while learning fine-grained retrieval without part supervision.}
    \label{fig:pif3d_overview}
\end{figure*}

\subsection{Overview}
\label{subsec:method_overview}

PIF-3D learns a shared retrieval space from shape-level triplets
$(\mathbf{X}_i,\mathcal{I}_i,c_i)$, where
$\mathbf{X}_i\in\mathbb{R}^{Q\times 6}$ is an XYZRGB point cloud,
$\mathcal{I}_i=\{\mathbf{I}_{i,n}\}_{n=1}^{N}$ contains $N$ rendered
views of the same shape, and $c_i$ is its natural-language caption.
The shape encoder maps the point cloud and renderings to a normalized
embedding $\mathbf{s}_i\in\mathbb{S}^{d-1}$, while the language encoder
maps the caption to $\mathbf{t}_i\in\mathbb{S}^{d-1}$.

The two shape modalities provide complementary evidence. Point-cloud
tokens encode three-dimensional geometry and surface color, whereas
multi-view dino.txt features provide language-aligned appearance and
local visual cues~\cite{jose2025dinov2}. PIF-3D summarizes the dense
patches within each view and jointly contextualizes the resulting view
tokens with local point-group tokens before global pooling. The model
is trained using only shape-level correspondences, without part labels,
segmentation masks, generated part captions, or auxiliary 3D--text
pretraining pairs.

\subsection{Point and Multi-View Tokens}
\label{subsec:tokenization}

\paragraph{Point-cloud tokens.}
A trainable PointBERT encoder~\cite{yu2022pointbert} partitions the
point cloud into local groups and produces $M$ group tokens:
\begin{equation}
    \mathbf{H}^{P}_i
    =
    E_P(\mathbf{X}_i)
    =
    \left[
        \mathbf{h}^{P}_{i,1};
        \ldots;
        \mathbf{h}^{P}_{i,M}
    \right]
    \in
    \mathbb{R}^{M\times d_P}.
\end{equation}
A trainable linear projection followed by layer normalization maps each
group token to the common fusion space:
\begin{align}
    \mathbf{p}_{i,m}
    &=
    \Pi_P\left(\mathbf{h}^{P}_{i,m}\right)
    =
    \operatorname{LN}
    \left(
        \mathbf{W}_P\mathbf{h}^{P}_{i,m}
        +
        \mathbf{b}_P
    \right),
    \nonumber\\
    \mathbf{P}_i
    &=
    \left[
        \mathbf{p}_{i,1};
        \ldots;
        \mathbf{p}_{i,M}
    \right]
    \in
    \mathbb{R}^{M\times d}.
    \label{eq:point_tokens}
\end{align}
PIF-3D uses these local group tokens rather than the global PointBERT
output, preserving spatially localized geometric and color evidence
for subsequent fusion.

\paragraph{Multi-view image tokens.}
For each rendering $\mathbf{I}_{i,n}$, $n\in\{1,\ldots,N\}$, the locked image encoder produces $K$ patch tokens:
\begin{equation}
    \mathbf{H}^{I}_{i,n}
    =
    E_I(\mathbf{I}_{i,n})
    =
    \left[
        \mathbf{h}^{I}_{i,n,1};
        \ldots;
        \mathbf{h}^{I}_{i,n,K}
    \right]
    \in
    \mathbb{R}^{K\times d_I}.
    \label{eq:image_patch_tokens}
\end{equation}
Across all views, these tokens form
\begin{equation}
    \mathbf{H}^{I}_i
=
\operatorname{stack}_{n=1}^{N}
\mathbf{H}^{I}_{i,n}
\in
\mathbb{R}^{N\times K\times d_I}.
    % \mathbf{H}^{I}_i
    % =
    % \left[
    %     \mathbf{H}^{I}_{i,1};
    %     \ldots;
    %     \mathbf{H}^{I}_{i,N}
    % \right]
    % \in
    % \mathbb{R}^{N\times K\times d_I}.
\end{equation}
The image-backbone parameters remain frozen. A trainable linear
projection followed by layer normalization is applied independently to
every patch token. The projected patches are then mean-pooled within
each view:
\begin{align}
    \mathbf{v}_{i,n}
    &=
    \frac{1}{K}
    \sum_{k=1}^{K}
    \Pi_I\left(\mathbf{h}^{I}_{i,n,k}\right),
    \nonumber\\
    \Pi_I\left(\mathbf{h}\right)
    &=
    \operatorname{LN}
    \left(
        \mathbf{W}_I\mathbf{h}
        +
        \mathbf{b}_I
    \right),
    \qquad n=1,\ldots,N.
    \label{eq:view_token}
\end{align}
This yields the multi-view token sequence
\begin{equation}
    \mathbf{V}_i
    =
    \left[
        \mathbf{v}_{i,1};
        \ldots;
        \mathbf{v}_{i,N}
    \right]
    \in
    \mathbb{R}^{N\times d}.
\end{equation}
Thus, the $N\times K$ patch-token grid is an intermediate
representation. The PIF-JT receives one pooled token for
each view rather than all $NK$ patch tokens. No additional learned view
or camera embeddings are used.

\subsection{PIF-Joint Transformer}
\label{subsec:pif_joint}

The PIF-Joint Transformer (PIF-JT) forms a single sequence by concatenating the $M$
projected point-group tokens with the $N$ pooled view tokens:
% \begin{equation}
%     \mathbf{Z}^{0}_i
%     =
%     \left[
%         \mathbf{P}_i;
%         \mathbf{V}_i
%     \right]
%     =
%     \left[
%         \mathbf{p}_{i,1};
%         \ldots;
%         \mathbf{p}_{i,M};
%         \mathbf{v}_{i,1};
%         \ldots;
%         \mathbf{v}_{i,N}
%     \right]
%     \in
%     \mathbb{R}^{(M+N)\times d}.
%     \label{eq:joint_sequence}
% \end{equation}
\begin{equation}
\begin{aligned}
\mathbf{Z}^{0}_i
&=
\left[\mathbf{P}_i;\mathbf{V}_i\right] \\
&=
\left[
\mathbf{p}_{i,1};\ldots;\mathbf{p}_{i,M};
\mathbf{v}_{i,1};\ldots;\mathbf{v}_{i,N}
\right] \\
&\in \mathbb{R}^{(M+N)\times d}.
\end{aligned}
\label{eq:joint_sequence}
\end{equation}

An $L$-layer PIF-JT contextualizes this sequence:
\begin{equation}
    \mathbf{Z}^{L}_i
    =
    F_{\mathrm{JT}}^{L}
    \left(
        \mathbf{Z}^{0}_i
    \right).
\end{equation}
Each layer is a pre-normalization Transformer encoder layer:
\begin{align}
    \mathbf{U}^{\ell}_i
    &=
    \mathbf{Z}^{\ell-1}_i
    +
    \operatorname{MSA}_{\ell}
    \left(
        \operatorname{LN}
        \left(
            \mathbf{Z}^{\ell-1}_i
        \right)
    \right),
    \nonumber\\
    \mathbf{Z}^{\ell}_i
    &=
    \mathbf{U}^{\ell}_i
    +
    \operatorname{MLP}_{\ell}
    \left(
        \operatorname{LN}
        \left(
            \mathbf{U}^{\ell}_i
        \right)
    \right),
    \qquad
    \ell=1,\ldots,L.
    \label{eq:pif_joint}
\end{align}
The feed-forward subnetwork maps each token from $d$ to $4d$ and back
to $d$. Because self-attention is evaluated over the complete
$(M+N)$-token sequence, every local point-group token can exchange
information with every rendered-view token. This distinguishes PIF-3D
from late fusion, where independently pooled modality representations
are combined only at the descriptor or similarity-score level.

Finally, a learned query $\mathbf{q}\in\mathbb{R}^{d}$ performs
multi-head attention pooling over the contextualized sequence:
\begin{align}
    \tilde{\mathbf{s}}_i
    &=
    \operatorname{MHA}
    \left(
        \mathbf{q},
        \operatorname{LN}(\mathbf{Z}^{L}_i),
        \operatorname{LN}(\mathbf{Z}^{L}_i)
    \right),
    \nonumber\\
    \mathbf{s}_i
    &=
    \frac{\tilde{\mathbf{s}}_i}
         {\left\|\tilde{\mathbf{s}}_i\right\|_2}.
    \label{eq:shape_pooling}
\end{align}
The learned query aggregates the $M+N$ contextualized tokens into a
single shape embedding while allowing their contributions to be
weighted adaptively.

\subsection{Locked Image Encoding and Text Adaptation}
\label{subsec:lora_adaptation}

PIF-3D builds on the pretrained dino.txt image--text space
~\cite{jose2025dinov2}. The image encoder and the original parameters
of the text encoder remain frozen. To adapt the language representation
to fine-grained 3D descriptions, we insert Low-Rank Adaptation
(LoRA)~\cite{hu2022lora} into linear transformations in the final
text-Transformer blocks. For a frozen linear transformation with weight
$\mathbf{W}_0$ and bias $\mathbf{b}_0$, the adapted output is
\begin{equation}
    \mathbf{y}
    =
    \mathbf{W}_0\mathbf{x}
    +
    \mathbf{b}_0
    +
    \frac{\alpha}{r}
    \mathbf{B}\mathbf{A}
    \operatorname{Dropout}(\mathbf{x}),
    \label{eq:lora}
\end{equation}
where $\mathbf{A}$ and $\mathbf{B}$ are trainable low-rank matrices,
$r$ is their rank, and $\alpha$ controls the update scale. The original
linear transformation remains fixed.

Let
$\mathbf{h}^{T}_i=E_T^{\mathrm{LoRA}}(c_i)$ denote the adapted global
caption representation. A trainable linear projection followed by
layer normalization maps it to the shared retrieval space:
\begin{align}
    \tilde{\mathbf{t}}_i
    &=
    \Pi_T\left(\mathbf{h}^{T}_i\right)
    =
    \operatorname{LN}
    \left(
        \mathbf{W}_T\mathbf{h}^{T}_i
        +
        \mathbf{b}_T
    \right),
    \nonumber\\
    \mathbf{t}_i
    &=
    \frac{\tilde{\mathbf{t}}_i}
         {\left\|\tilde{\mathbf{t}}_i\right\|_2}.
    \label{eq:text_embedding}
\end{align}
Consequently, PointBERT, the point, image, and text projections, the
PIF-JT, the text-side LoRA parameters, and the contrastive
logit scale are optimized for retrieval. The pretrained dino.txt image
encoder and original text-encoder parameters remain fixed.

\subsection{Symmetric Shape--Text Alignment}
\label{subsec:contrastive_alignment}

For a mini-batch of $B$ matched shape--caption pairs, the logit between
shape $i$ and caption $j$ is
\begin{equation}
    \ell_{ij}
    =
    \gamma\,
    \mathbf{s}_i^{\top}\mathbf{t}_j,
    \qquad
    \gamma
    =
    \min\left\{\exp(\eta),100\right\},
    \label{eq:similarity}
\end{equation}
where $\eta$ is a learnable scalar and $\gamma$ is initialized to $10$.
Following CLIP-style contrastive learning~\cite{radford2021learning},
we optimize both retrieval directions:
\begin{align}
    \mathcal{L}_{S\rightarrow T}
    &=
    -\frac{1}{B}
    \sum_{i=1}^{B}
    \log
    \frac{\exp(\ell_{ii})}
         {\sum_{j=1}^{B}\exp(\ell_{ij})},
    \nonumber\\
    \mathcal{L}_{T\rightarrow S}
    &=
    -\frac{1}{B}
    \sum_{i=1}^{B}
    \log
    \frac{\exp(\ell_{ii})}
         {\sum_{j=1}^{B}\exp(\ell_{ji})},
    \nonumber\\
    \mathcal{L}
    &=
    \frac{1}{2}
    \left(
        \mathcal{L}_{S\rightarrow T}
        +
        \mathcal{L}_{T\rightarrow S}
    \right).
    \label{eq:symmetric_loss}
\end{align}
Matched pairs form the diagonal positives, while all remaining
shape--caption combinations in the mini-batch serve as negatives.

At inference, each gallery shape is encoded once from its point cloud
and rendered views. A caption query is encoded independently, and
gallery shapes are ranked by $\mathbf{t}^{\top}\mathbf{s}_j$. Since
both embeddings are $\ell_2$-normalized, this score is cosine
similarity. Shape-to-text retrieval uses the same embedding space with
the similarity matrix transposed.

%%%%%%%%%%%%%%%%%%%%%%%%%%%%%%%%%%%%%%%%%%%%%%%%%%%% Experiments %%%%%%%%%%%%%%%%%%%%%%%%%%%%%%%%%%%%%%%%%%%%%%%%%%%%
\section{Experiments}
\label{sec:experiments}
\subsection{Text-to-3D Retrieval on Text2Shape}
\label{subsec:text2shape}

\paragraph{Benchmark:} Text2Shape~\cite{chen2018text2shape} is the standard benchmark for fine-grained text-to-3D retrieval. It pairs human-written descriptions with colored ShapeNet chairs and tables. The captions describe detailed geometric and appearance properties, including shape, color, material, texture, and local object components. We evaluate both shape-to-text (S2T) and text-to-shape (T2S) retrieval using Recall Rate at ranks one and five (RR@1 and RR@5) and Normalized Discounted Cumulative Gain at rank five (NDCG@5).

\paragraph{Experimental Setup:}
We follow the standard Text2Shape data and evaluation protocol used in prior work~\cite{chen2018text2shape,ruan2024tricolo,ren2025sca3d}, including the provided instance-disjoint training, validation, and test splits. Each training sample consists of a colored point cloud, multi-view renderings of the same shape, and one of its benchmark-provided captions. We use the benchmark provided $2{,}500$ colored point cloud and six rendered views per shape, details are provided in the supplementary material. Training views are sampled randomly, whereas evaluation views are selected deterministically and evenly from the available renderings.

\paragraph{Implementation Details:}
Our reference PIF-3D configuration uses a 512-dimensional joint embedding space, concatenation fusion with attention pooling, one self-attention layer, eight attention heads, and no view embeddings. PointBERT is trained jointly, while the dino.txt image encoder remains frozen and rank-8 LoRA is applied to the final 24 text layers. We optimize the symmetric
contrastive objective using AdamW for 20 epochs with a batch size of 256, learning rates of $10^{-4}$ and $10^{-5}$ for the main model and LoRA parameters, respectively, and cosine decay with a 2\% warm-up. We report the mean and sample standard deviation over three independent runs with seeds $\{0,21,42\}$. All experiments are tracked using Weights \& Biases~\cite{wandb}. Additional details on implementation, computing infrastructure, and reproducibility are provided in the supplementary material.

\paragraph{Results:}
As shown in Table~\ref{tab:text2shape_results}, PIF-3D establishes a new state of the art on Text2Shape, achieving the best performance across all six metrics. It surpasses diverse prior approaches, including view-sequence modeling with Y2Seq2Seq~\cite{han2019y2seq2seq}, trimodal contrastive learning with TriCoLo~\cite{ruan2024tricolo}, part--word matching with Parts2Words~\cite{tang2023parts2words}, cross-view correspondence with COM3D~\cite{wu2024com3d}, component-level data augmentation with SCA3D~\cite{ren2025sca3d}, and large-scale 3D pretraining with Uni3D and 3DAlign-DAER~\cite{zhou2024uni3d,fan20263dalign}. PIF-3D achieves $32.16\pm1.09$ S2T RR@1 and $18.86\pm0.19$ T2S RR@1. Compared with the strongest prior method, 3DAlign-DAER, it improves S2T RR@1 by $4.05$ percentage points and T2S RR@1 by $1.33$ points, while also improving the corresponding RR@5 and NDCG@5 results. Notably, this state-of-the-art
performance is achieved without semantic part-segmentation labels, generated part captions, augmented 3D--text pairs, or additional 3D--text pretraining data.

\begin{table*}[t]
\centering
{\small
\setlength{\tabcolsep}{5pt}
\begin{tabular}{llrrrrrr}
\hline
Method & Venue & S2T RR@1 & S2T RR@5 & S2T NDCG@5 & T2S RR@1 & T2S RR@5 & T2S NDCG@5 \\
\hline
Text2Shape   & ACCV'18 & 0.83  & 3.37  & 0.73  & 0.40  & 2.37  & 1.35  \\
Y2Seq2Seq    & AAAI'19 & 6.77  & 19.30 & 5.30  & 2.93  & 9.23  & 6.05  \\
TriCoLo      & WACV'24 & 16.33 & 45.52 & 12.73 & 10.25 & 29.07 & 19.85 \\
Parts2Words\textsuperscript{\dag}
             & CVPR'23 & 19.38 & 47.17 & 15.30 & 12.72 & 32.98 & 23.13 \\
COM3D\textsuperscript{\dag}       
             & ICME'24 & 20.03 & 48.32 & 15.62 & 13.12 & 33.48 & 23.89 \\
Uni3D-B\textsuperscript{\ddag}
             & ICLR'24 & 22.51 & 51.39 & 17.05 & 14.57 & 36.28 & 25.04 \\
Uni3D-L\textsuperscript{\ddag}
             & ICLR'24 & 25.02 & 53.55 & 18.71 & 15.21 & 37.57 & 27.13 \\
Uni3D-g\textsuperscript{\ddag}
             & ICLR'24 & 26.23 & 54.71 & 19.42 & 16.51 & 38.32 & 28.06 \\
SCA3D\textsuperscript{\dag,\ddag}
             & ICRA'25 & 27.22 & 55.56 & 19.04 & 16.67 & 38.90 & 28.17 \\
3DAlign-DAER\textsuperscript{\ddag}
             & AAAI'26 & 28.11 & 56.58 & 19.93 & 17.53 & 39.88 & 29.06 \\
PIF-3D       & -- & $\mathbf{32.16\pm 1.09}$ & $\mathbf{60.04 \pm 0.48}$ & $\mathbf{22.74 \pm 0.21}$ &
                 $\mathbf{18.86 \pm 0.19}$ & $\mathbf{41.62 \pm 0.24}$ & $\mathbf{30.72 \pm 0.16}$ \\
\hline
\end{tabular}
}
% \caption{Results on the Text2Shape benchmark.\textsuperscript{\dag} denotes methods using semantic part-segmentation annotations.\textsuperscript{\ddag} denotes methods using auxiliary 3D--text training data: Uni3D is pretrained on nearly 1M 3D shapes paired with 70M texts, SCA3D constructs augmented pairs using generated part captions, and 3DAlign-DAER trains on Align3D-2M, containing approximately 2M text--3D pairs. PIF-3D uses neither semantic part annotations nor auxiliary 3D--text data, yet achieves SOTA performance across all metrics. The Uni3D results are reported by 3DAlign-DAER~\cite{fan20263dalign} for models introduced in Uni3D~\cite{zhou2024uni3d}. PIF-3D results are reported as mean $\pm$ sample standard deviation over three seeds.}

\caption{Comparison with prior work on Text2Shape. \textsuperscript{\dag} indicates methods using semantic part-segmentation supervision; \textsuperscript{\ddag} indicates methods using auxiliary 3D--text data. PIF-3D establishes a new state of the art across all six metrics without either resource. Uni3D values are those reported by 3DAlign-DAER~\cite{fan20263dalign}; PIF-3D values are mean $\pm$ sample standard deviation over three seeds.}
\label{tab:text2shape_results}
\end{table*}

%%%%%%%%%%%%%%%%%%%%%%%%%%%%%%%%%%%%%%%%%%%%%%%%%%%% Results on 3DCoMPaT200 %%%%%%%%%%%%%%%%%%%%%%%%%%%%%%%%%%%%%%%%%%%%%%%%%%%%
\subsection{Compositional Part-Based Retrieval on 3DCoMPaT200}
\label{subsec:compat200}

\paragraph{Benchmark:} 3DCoMPaT200~\cite{ahmed20253dcompat200} introduces compositional shape retrieval, in which a text query specifies one, three, or six part--material--color constraints and the model retrieves the corresponding stylized 3D shape. We evaluate text-to-shape retrieval using R@1 and R@5. Because the original protocol uses 8,196-point colored clouds, whereas the public HDF5 files contain only 2,048-point colored clouds, we conduct a controlled public-release adaptation rather than reproduce the originally reported results.

\paragraph{Experimental Setup:} We retrain every method using the same filtered public training data and the provided six-part captions. Evaluation uses the official one-, three-, and six-part validation queries with their corresponding shape galleries, with disjoint training and evaluation shapes. Full details of the filtering, split construction, and relevance definition are provided in the supplementary material.

\paragraph{Implementation Details:}
PIF-3D directly transfers the architecture, six-view configuration, loss, learning rates, and scheduler selected on Text2Shape, without 3DCoMPaT200-specific hyperparameter tuning. The only input-level change is the use of the publicly available 2,048-point clouds. ULIP and OpenShape retain the $5\times10^{-4}$ learning rate from their public training recipes. For a controlled and fair comparison, all methods are trained on the same data with a global batch size of 200 and exactly 7,170 optimizer updates. We report results from the predetermined final epoch as mean $\pm$ sample standard deviation over seeds $\{0,21,42\}$, avoiding checkpoint selection on the reported one-, three-, and six-part validation splits.

\begin{table*}[t]
\centering
\small
\setlength{\tabcolsep}{4.5pt}
\begin{tabular}{@{}lcccccc@{}}
\hline
Method
& \multicolumn{2}{c}{1 Part}
& \multicolumn{2}{c}{3 Parts}
& \multicolumn{2}{c}{6 Parts} \\
\cline{2-7}
& R@1 & R@5
& R@1 & R@5
& R@1 & R@5 \\
\hline
ULIP
& $8.53\pm0.80$
& $34.91\pm0.52$
& $17.53\pm0.48$
& $62.49\pm1.07$
& $21.45\pm1.01$
& $72.26\pm0.98$ \\

OpenShape
& $8.81\pm0.36$
& $36.54\pm0.30$
& $14.54\pm0.54$
& $57.11\pm0.30$
& $17.02\pm0.56$
& $63.97\pm0.25$ \\
\hline
ULIP + LF-1V
& $10.82\pm0.12$
& $\mathbf{46.20}\pm0.53$
& $15.29\pm0.25$
& $58.33\pm0.72$
& $17.24\pm0.40$
& $64.09\pm0.65$ \\

OpenShape + LF-1V
& $\mathbf{11.05}\pm0.21$
& $45.63\pm0.77$
& $13.83\pm0.33$
& $55.63\pm0.74$
& $15.64\pm0.92$
& $60.54\pm1.15$ \\
\hline
PIF-3D (frozen language)
& $9.00\pm0.68$
& $38.08\pm0.77$
& $17.63\pm1.06$
& $\mathbf{64.87}\pm0.60$
& $24.23\pm0.63$
& $77.74\pm0.60$ \\

% (Text LoRA, 6V)
PIF-3D
& $9.14\pm0.24$
& $37.11\pm0.77$
& $\mathbf{17.67}\pm0.48$
& $64.66\pm0.33$
& $\mathbf{26.50}\pm0.28$
& $\mathbf{79.75}\pm0.52$ \\
\hline
\end{tabular}
% \caption{Text-to-shape retrieval on the 3DCoMPaT200 public-release 2,048-point protocol. Results are percentages reported as mean $\pm$ sample standard deviation over three seeds. LF-1V averages point-cloud and single-view similarities, whereas PIF-3D jointly processes six views. Both PIF-3D variants use the same architectural setup as the one used on Text2Shapes.}
\caption{Text-to-shape retrieval under our controlled 2,048-point public-release protocol for 3DCoMPaT200. LF-1V averages point-cloud and single-view similarities, whereas PIF-3D jointly fuses point tokens with six views. PIF-3D's advantage grows with query complexity and is largest for six-part queries. Values are percentages reported as mean $\pm$ sample standard deviation over three seeds.}
\label{tab:compat200_results}
\end{table*}

\paragraph{Results and Compositionality Analysis:} Table~\ref{tab:compat200_results} reveals a clear dependence on query complexity. Single-view late fusion performs best for one-part queries, where OpenShape~+~LF-1V achieves the highest R@1 and ULIP~+~LF-1V the highest R@5. At three parts, the PIF-3D variants obtain the best R@1 and R@5, although the improvements over ULIP remain modest. The advantage becomes substantially larger for six-part queries. PIF-3D achieves $26.50\pm0.28$ R@1 and $79.75\pm0.52$ R@5, improving the strongest non-PIF method, ULIP, by $5.05$ and $7.49$ percentage points, respectively. The frozen-language variant already exceeds ULIP by $2.78$ R@1 and $5.48$ R@5, showing that the improvement does not arise solely from language adaptation. Text LoRA further improves the frozen model by $2.27$ R@1 and $2.01$ R@5 at six parts. These results indicate that the benefit of joint point--image fusion grows as retrieval requires satisfying more simultaneous part-level constraints. Because LF-1V uses one rendering whereas PIF-3D uses six, we treat it as a simple render-assisted reference rather than as a view-matched baseline.

%%%%%%%%%%%%%%%%%%%%%%%%%%%%%%%%%%%%%%%%%%%%%%%%%%%% Ablation Studies %%%%%%%%%%%%%%%%%%%%%%%%%%%%%%%%%%%%%%%%%%%%%%%%%%%%
\subsection{Ablation Studies}
\label{subsec:ablations}
We conduct controlled ablations on Text2Shape to isolate the contributions of the model components and key design choices. Unless explicitly varied, each experiment uses the reference configuration described above, including $2{,}500$ colored points, six rendered views, a trainable scale-1 PointBERT encoder, a frozen dino.txt image encoder, and symmetric contrastive learning. We change only the factor under investigation while retaining the training schedule, batch size, and data-sampling policy. For each configuration, we select the checkpoint using the validation mean of S2T and T2S RR@1 and evaluate the test set once. Results are reported as mean $\pm$ sample standard deviation over seeds $\{0,21,42\}$.

%%%%%%%%%%%%%%%%%%%%%%%%%%%%%%% Ablation 1 : Contribution of the input modalities %%%%%%%%%%%%%%%%%%%%%%%%%%%%%%%
\paragraph{Contribution of the input modalities.} Table~\ref{tab:t2s_component_ablation} separates the contributions of geometry and rendered appearance. The image-only model substantially outperforms the point-only model, indicating that the language-aligned render features provide the stronger individual representation. Point clouds nevertheless contribute complementary evidence: averaging the independently trained point-only and image-only similarities improves the image-only model by $2.15$ S2T RR@1 and $0.95$ T2S RR@1. Joint token fusion yields further gains of $2.60$ and $0.97$ points over late fusion on S2T and T2S RR@1, respectively, while also improving both RR@5 scores. Since the latter two models use the same input modalities, these gains isolate the benefit of learning point--image interactions before pooling instead of combining independently produced retrieval scores.

\begin{table}[t]
\centering
{\small
\setlength{\tabcolsep}{1mm}
\begin{tabular}{@{}llcc@{}}
\hline
Method & Dir. & RR@1 & RR@5 \\
\hline
Point only
  & S2T & $17.26\pm1.42$ & $42.18\pm1.20$ \\
  & T2S & $10.90\pm0.24$ & $30.00\pm0.17$ \\
\hline
Image only
  & S2T & $27.41\pm0.70$ & $54.99\pm0.22$ \\
  & T2S & $16.94\pm0.18$ & $37.87\pm0.31$ \\
\hline
Late fusion
  & S2T & $29.56\pm1.31$ & $58.41\pm0.80$ \\
  & T2S & $17.89\pm0.19$ & $40.33\pm0.39$ \\
\hline
PIF-3D
  & S2T & $\mathbf{32.16}\pm1.09$ & $\mathbf{60.04}\pm0.48$ \\
  & T2S & $\mathbf{18.86}\pm0.19$ & $\mathbf{41.62}\pm0.24$ \\
\hline
\end{tabular}
}
% \caption{Contribution of point-cloud and rendered-image information on
% Text2Shape. Late fusion averages the similarities produced by independently
% trained point-only and image-only models using fixed weights of $0.5/0.5$.
% PIF-3D instead learns a joint point--image representation. Results are
% test-set metrics obtained from validation-selected checkpoints and are
% reported as mean $\pm$ sample standard deviation over three seeds.}
\caption{Input-modality ablation on Text2Shape. Late fusion averages similarities from separately trained point-only and image-only models with fixed weights of $0.5/0.5$, whereas PIF-3D learns point--image interactions before pooling. Its consistent improvement over late fusion isolates the benefit of joint token fusion.}
\label{tab:t2s_component_ablation}
\end{table}

%%%%%%%%%%%%%%%%%%%%%%%%%%%%%%% Ablation 2 : DINO.txt Adaptation %%%%%%%%%%%%%%%%%%%%%%%%%%%%%%%
\paragraph{Vision-language encoder adaptation.} Table~\ref{tab:t2s_dino_ablation} examines which dino.txt branch to adapt and how much language-side capacity is required. At the same two-layer depth, text LoRA performs best on three of the four metrics; image LoRA is highest only on S2T RR@1, while adapting both branches provides no additional benefit. This supports retaining the pretrained visual representation and specializing the language interface. Adapting only the final two or four text layers produces similar results, whereas performance improves consistently from eight to 24 adapted layers. The final 24-layer configuration improves over frozen dino.txt by $3.25$ S2T RR@1, $4.13$ S2T RR@5, $1.93$ T2S RR@1, and $3.02$ T2S RR@5. Thus, fine-grained 3D captions benefit from broad parameter-efficient adaptation of the text tower without modifying the locked image encoder.

\begin{table*}[t]
\centering
\small
\setlength{\tabcolsep}{5pt}
\begin{tabular}{@{}llccccc@{}}
\hline
Setting & Updated branch & Layers
& \multicolumn{2}{c}{S2T}
& \multicolumn{2}{c}{T2S} \\
\cline{4-7}
& & & RR@1 & RR@5 & RR@1 & RR@5 \\
\hline
\multicolumn{7}{l}{\emph{Adaptation target at a matched depth}} \\
Frozen dino.txt       & None         & 0      & $28.91\pm0.37$ & $55.91\pm0.56$ & $16.93\pm0.27$ & $38.60\pm0.15$ \\
Image LoRA            & Image        & Last 2 & $\mathbf{29.52}\pm0.90$ & $56.12\pm0.58$ & $17.11\pm0.35$ & $38.92\pm0.24$ \\
Text LoRA             & Text         & Last 2 & $29.14\pm0.38$ & $\mathbf{57.06}\pm0.98$ & $\mathbf{17.41}\pm0.45$ & $\mathbf{39.43}\pm0.32$ \\
Image + text LoRA     & Image + text & Last 2 & $29.32\pm0.37$ & $56.90\pm0.71$ & $17.22\pm0.26$ & $39.27\pm0.23$ \\
% Image + text LoRA     & Image + text & Last 4 & $28.96\pm0.38$ & $57.91\pm1.14$ & $17.53\pm0.39$ & $39.88\pm0.25$ \\
\hline
\multicolumn{7}{l}{\emph{Text-LoRA depth}} \\
Text LoRA             & Text         & Last 2  & $29.14\pm0.38$ & $57.06\pm0.98$ & $17.41\pm0.45$ & $39.43\pm0.32$ \\
Text LoRA             & Text         & Last 4  & $29.09\pm0.42$ & $56.88\pm0.38$ & $17.37\pm0.22$ & $39.34\pm0.08$ \\
Text LoRA             & Text         & Last 8  & $29.97\pm0.07$ & $58.11\pm0.68$ & $18.04\pm0.07$ & $40.08\pm0.15$ \\
Text LoRA             & Text         & Last 16 & $30.55\pm0.69$ & $58.90\pm0.30$ & $18.51\pm0.28$ & $40.61\pm0.13$ \\
Text LoRA             & Text         & Last 24 & $\mathbf{32.16\pm 1.09}$ & $\mathbf{60.04 \pm 0.48}$ & $\mathbf{18.86 \pm 0.19}$ & $\mathbf{41.62 \pm 0.24}$ \\
\hline
\end{tabular}
\caption{Vision--language encoder adaptation on Text2Shape. LoRA targets are compared at a matched two-layer depth, while the depth study applies rank-8 LoRA only to the text tower. Text-only adaptation is strongest overall, and extending it to all 24 layers yields the best results while keeping the image encoder locked. The Last-2 text result is repeated to anchor both comparisons.}
\label{tab:t2s_dino_ablation}
\end{table*}

%%%%%%%%%%%%%%%%%%%%%%%%%%%%%%% Ablation 3 : Point--image fusion design on Text2Shape %%%%%%%%%%%%%%%%%%%%%%%%%%%%%%%
\paragraph{Point--image fusion design.}
Table~\ref{tab:t2s_fusion_ablation} compares joint self-attention with directional and bidirectional cross-attention. Concatenation followed by self-attention achieves the strongest S2T results and best average performance. Updating image tokens from point tokens improves T2S by only $0.21$ RR@1 and $0.10$ RR@5 but reduces S2T performance. Neither the reverse direction nor bidirectional cross-attention provides consistent gains, suggesting that joint token contextualization matters more than attention direction. We therefore use the simpler and more balanced concatenation design in PIF-3D.

\begin{table*}[t]
\centering
\small
\setlength{\tabcolsep}{5pt}
\begin{tabular}{@{}lccccc@{}}
\hline
Fusion design & Attn. layers
& \multicolumn{2}{c}{S2T}
& \multicolumn{2}{c}{T2S} \\
\cline{3-6}
& & RR@1 & RR@5 & RR@1 & RR@5 \\
\hline
Concatenation + self-attention & 1 & $\mathbf{32.16}\pm1.09$ & $\mathbf{60.04}\pm0.48$ & $18.86\pm0.19$ & $41.62\pm0.24$ \\
Point $\leftarrow$ Image       & 1 & $31.38\pm1.72$ & $58.97\pm0.58$ & $18.63\pm0.14$ & $41.13\pm0.35$ \\
Image $\leftarrow$ Point       & 1 & $31.85\pm1.47$ & $59.57\pm0.57$ & $\mathbf{19.07}\pm0.17$ & $\mathbf{41.72}\pm0.30$ \\
Bidirectional                  & 1 & $30.53\pm0.49$ & $58.56\pm0.81$ & $18.41\pm0.07$ & $41.22\pm0.43$ \\
\hline
\end{tabular}
% \caption{Point--image fusion design on Text2Shape. Point $\leftarrow$ Image denotes point queries attending to image keys and values; Image $\leftarrow$ Point denotes the reverse. All configurations use the same token dimension, number of heads, six views, and one-layer attention encoder.}
\caption{Point--image fusion ablation on Text2Shape. $A\leftarrow B$ denotes queries from $A$ attending to keys and values from $B$. Concatenation with joint self-attention provides the strongest bidirectional balance, indicating that shared contextualization matters more than prescribing a cross-attention direction.}
\label{tab:t2s_fusion_ablation}
\end{table*}

%%%%%%%%%%%%%%%%%%%%%%%%%%%%%%% Ablation 4 : Sensitivity to the number of rendered views %%%%%%%%%%%%%%%%%%%%%%%%%%%%%%%
\paragraph{Sensitivity to the number of rendered views.}
As Table~\ref{tab:t2s_view_count} shows, increasing from one to six views improves S2T RR@1/RR@5 by $12.77/14.70$ points and T2S RR@1/RR@5 by $5.86/8.57$ points. Beyond six views, gains are small and non-monotonic: ten views improve these metrics by only $0.23$--$0.72$ points, while 12 provide no consistent benefit. We therefore use six views throughout, capturing most gains with four fewer rendered inputs than the ten-view variant.

\begin{table}[t]
\centering
{\small
\setlength{\tabcolsep}{1mm}
\begin{tabular}{@{}llcc@{}}
\hline
Views & Dir. & RR@1 & RR@5 \\
\hline
1
  & S2T & $19.39\pm0.84$ & $45.34\pm1.19$ \\
  & T2S & $13.00\pm0.49$ & $33.05\pm0.59$ \\
\hline
3
  & S2T & $27.18\pm0.61$ & $55.71\pm0.66$ \\
  & T2S & $17.09\pm0.18$ & $38.83\pm0.16$ \\
\hline
6
  & S2T & $32.16\pm1.09$ & $60.04\pm0.48$ \\
  & T2S & $18.86\pm0.19$ & $41.62\pm0.24$ \\
\hline
8
  & S2T & $31.83\pm0.54$ & $60.25\pm0.81$ \\
  & T2S & $19.38\pm0.40$ & $41.78\pm0.18$ \\
\hline
10
  & S2T & $\mathbf{32.39}\pm0.18$ & $\mathbf{60.76}\pm0.10$ \\
  & T2S & $19.38\pm0.45$ & $\mathbf{42.30}\pm0.48$ \\
\hline
12
  & S2T & $32.12\pm0.60$ & $60.18\pm0.33$ \\
  & T2S & $\mathbf{19.41}\pm0.28$ & $42.28\pm0.41$ \\
\hline
\end{tabular}
}
\caption{Sensitivity to the number of rendered views on Text2Shape, using the same view count during training and evaluation. Performance rises sharply through six views and then largely saturates, supporting six views as the reference accuracy--efficiency trade-off.}
\label{tab:t2s_view_count}
\end{table}

%%%%%%%%%%%%%%%%%%%%%%%%%%%%%%% Qualitative Results and Limitations %%%%%%%%%%%%%%%%%%%%%%%%%%%%%%%
\subsection{Qualitative Results and Limitations}
\begin{figure}[t]
    \centering
    \includegraphics[width=\columnwidth]{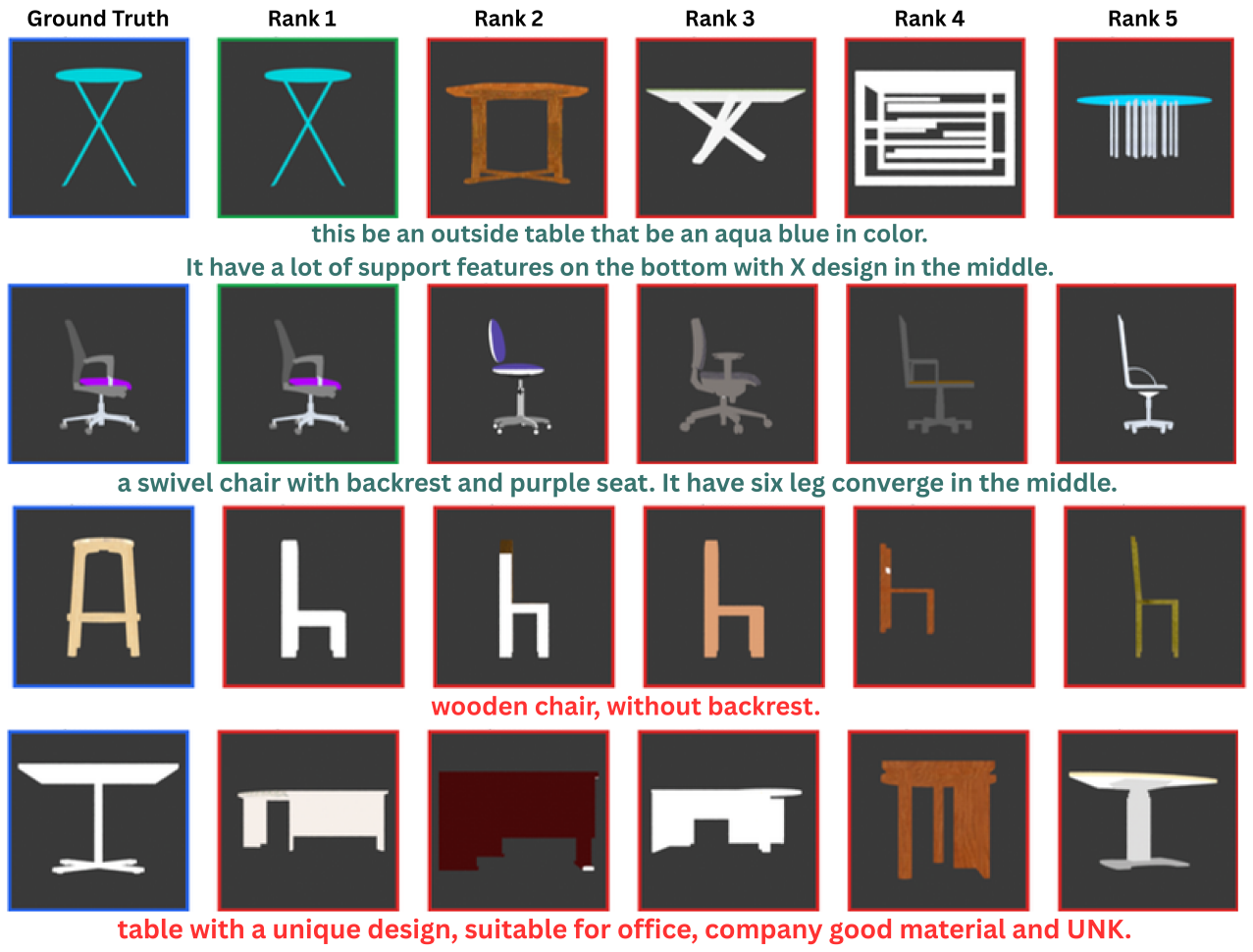}
    \caption{Text2Shape retrievals: the first two queries succeed at rank 1, while the last two fail within the top five. Boxes in blue marks targets; green/red mark correct/incorrect results.}
    \label{fig:qualitative_results}
\end{figure}
Figure~\ref{fig:qualitative_results} shows that PIF-3D can jointly match distinctive color and geometry, such as the aqua X-supported table and purple swivel chair. Failures occur for negation (``without backrest'') and vague or noisy descriptions containing terms such as ``unique design'' and ``UNK,'' revealing sensitivity to absent attributes and weak textual cues. Additional qualitative results can be found in the supplementary material.

\section{Conclusion}
We presented PIF-3D, a segmentation-free point-image fusion framework for fine-grained text-to-3D retrieval. Without part annotations or auxiliary 3D--text data, it achieves state-of-the-art Text2Shape results and strong gains on compositional 3DCoMPaT200 queries. Ablations show that joint fusion outperforms both unimodal variants.

%%%%%%%%%%%%%%%%%%%%%%%%%%%%%%%%%%%%%%%%%%%%%%%%%%%% References (Bibliography) %%%%%%%%%%%%%%%%%%%%%%%%%%%%%%%%%%%%%%%%%%%%%%%%%%%%
\bibliography{aaai2027}


\end{document}